\documentclass[letterpaper, 10 pt, journal, twoside]{IEEEtran}
\usepackage{amsmath,amsfonts}
\usepackage{algorithmic}
\usepackage{algorithm}
\usepackage{array}
\usepackage[caption=false,font=normalsize,labelfont=sf,textfont=sf]{subfig}
\usepackage{textcomp}
\usepackage{stfloats}
\usepackage{url}
\usepackage{verbatim}
\usepackage{graphicx}
\usepackage{cite}
\usepackage{makecell}
\usepackage{booktabs}
\usepackage{arydshln} 
\usepackage{multirow}
\usepackage{balance}
\usepackage{fancyhdr}
\usepackage{amssymb}
\usepackage{color}
\usepackage{array}
\usepackage[switch]{lineno}
\usepackage{bbding}
\usepackage{tabularx}
\usepackage{hyperref}

\usepackage{tikz,xcolor,hyperref}
\definecolor{lime}{HTML}{A6CE39}
\DeclareRobustCommand{\orcidicon}{%
    \begin{tikzpicture}
    \draw[lime, fill=lime] (0,0) 
    circle [radius=0.16] 
    node[white] {
   {\fontfamily{qag}\selectfont \tiny ID}};    \draw[white, fill=white] (-0.0625,0.095) 
    circle [radius=0.007];    \end{tikzpicture}
    \hspace{-2mm}}
\foreach \x in {A, ..., Z}{%
    \expandafter\xdef\csname orcid\x\endcsname{\noexpand\href{https://orcid.org/\csname orcidauthor\x\endcsname}{\noexpand\orcidicon}}
    }

\hypersetup{colorlinks = true, 
	    linkcolor = red,
	    urlcolor = blue,
            citecolor = green}

\begin{document}


\title{HE-Nav: A High-Performance and Efficient Navigation System for Aerial-Ground Robots in Cluttered Environments}

\author{Junming Wang\orcidA{}, Zekai Sun\orcidB{}, Xiuxian Guan\orcidC{}, Tianxiang Shen\orcidD{}, Dong Huang, Zongyuan Zhang,\\ Tianyang Duan, Fangming Liu\orcidE{},~\IEEEmembership{Senior Member,~IEEE} and Heming Cui\orcidF{},~\IEEEmembership{Member,~IEEE}

\thanks{Manuscript received: April 18, 2024; Revised: July 1, 2024; Accepted: September 13, 2024. This paper was recommended for publication by Editor Markus Vincze upon evaluation of the Associate Editor and Reviewers’ comments. \textit{(Corresponding author: Heming Cui.)} 

Junming Wang, Zekai Sun, Xiuxian Guan, Tianxiang Shen, Dong Huang, Zongyuan Zhang, Tianyang Duan are with the University of Hong Kong, Hong Kong SAR, China. (e-mail: jmwang@cs.hku.hk). 

Heming Cui is with the University of Hong Kong, Hong Kong SAR, China, and also with the the Shanghai AI Labolatory, China. (e-mail: heming@cs.hku.hk). 

Fangming Liu is with Peng Cheng Laboratory, and Huazhong University of Science and Technology, China. (e-mail: fmliu@hust.edu.cn). 
}
\thanks{Digital Object Identifier (DOI): see top of this page.}}

\markboth{IEEE ROBOTICS AND AUTOMATION LETTERS. PREPRINT VERSION. ACCEPTED SEPTEMBER, 2024}
{Wang \MakeLowercase{\textit{et al.}}: HE-Nav: A High-Performance and Efficient Navigation System for Aerial-Ground Robots in Occluded Environments}


\maketitle

\begin{abstract}
Existing AGR navigation systems have advanced in lightly occluded scenarios (e.g., buildings) by employing 3D semantic scene completion networks for voxel occupancy prediction and constructing Euclidean Signed Distance Field (ESDF) maps for collision-free path planning. However, these systems exhibit suboptimal performance and efficiency in cluttered environments with severe occlusions (e.g., dense forests or tall walls), due to limitations arising from perception networks' low prediction accuracy and path planners' high computational overhead.

In this paper, we present HE-Nav, the first high-performance and efficient navigation system tailored for AGRs operating in cluttered environments. The perception module utilizes a lightweight semantic scene completion network (LBSCNet), guided by a bird's eye view (BEV) feature fusion and enhanced by an exquisitely designed SCB-Fusion module and attention mechanism. This enables real-time and efficient obstacle prediction in cluttered areas, generating a complete local map. Building upon this completed map, our novel AG-Planner employs the energy-efficient kinodynamic A* search algorithm to guarantee planning is energy-saving. Subsequent trajectory optimization processes yield safe, smooth, dynamically feasible and ESDF-free aerial-ground hybrid paths. Extensive experiments demonstrate that HE-Nav achieved 7x energy savings in real-world situations while maintaining planning success rates of 98\% in simulation scenarios. Code and video are available on our project page: \url{https://jmwang0117.github.io/HE-Nav/}.
\end{abstract}


\section{INTRODUCTION}
\begin{figure}[t] 
  \centering
  \includegraphics[width=\linewidth]{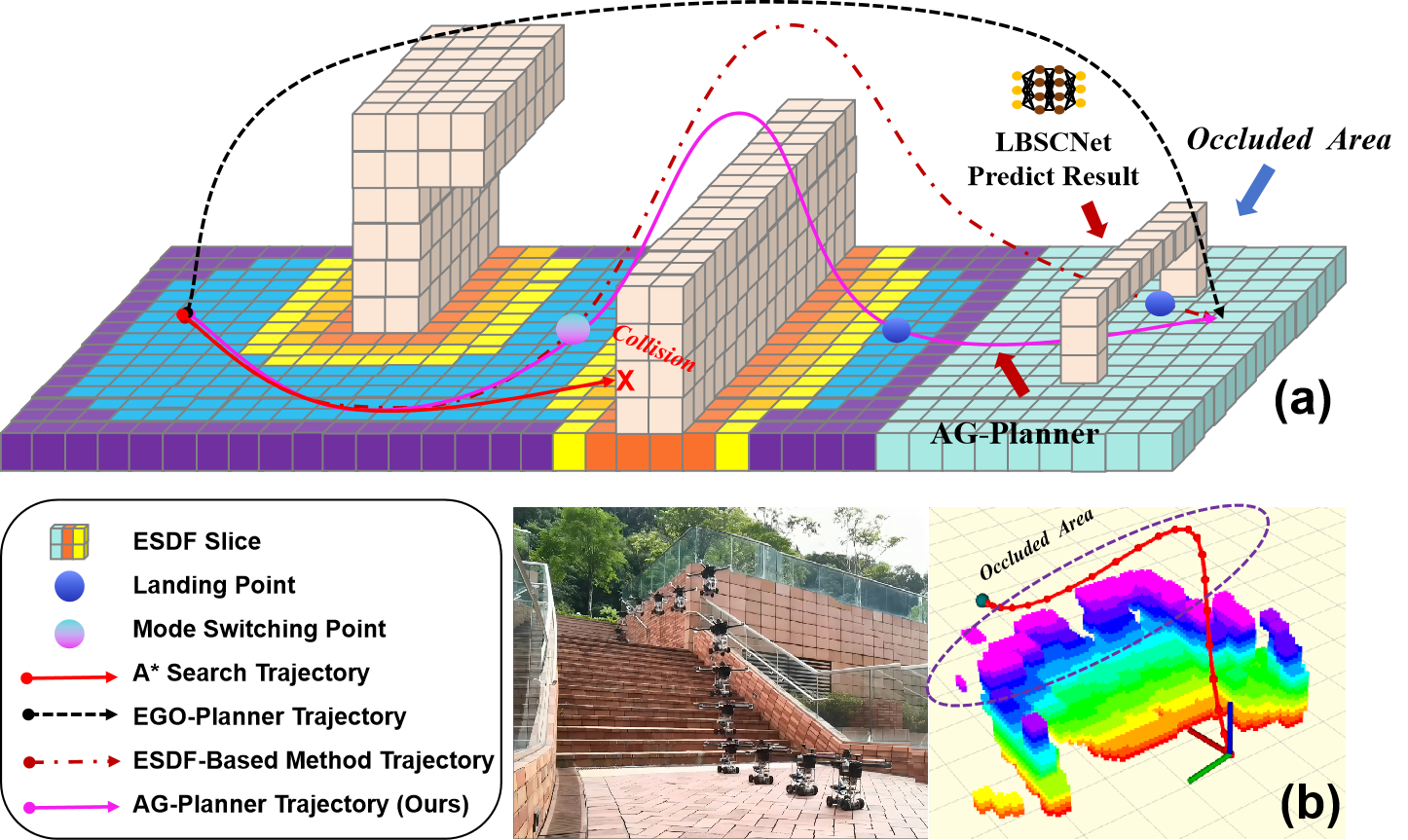}
  \caption{(a) Current navigation systems underperform in occluded areas due to inaccurate obstacle prediction and the computationally intensive process of creating ESDF maps. (b) Our HE-Nav system can generate energy-saving, collision-free and ESDF-free aerial-ground paths in real-time with the help of the LBSCNet model and AG-Planner. }
  \label{fig:head}
\end{figure}

\IEEEPARstart{I}{n} recent years, aerial-ground robots (AGRs) \cite{fan2019autonomous,zhang2022autonomous, jmwang} have emerged as a promising solution for search and rescue tasks \cite{tan2021multimodal}. This is attributed to their exceptional mobility and long endurance, which enable them to seamlessly switch between aerial and ground modes, allowing for hybrid locomotion (i.e., flying and driving) in the above challenging tasks. Specifically, the \textit{perception module} and \textit{path planner} are two key components in the AGRs navigation system \cite{zhang2022autonomous,fan2019autonomous}, working together to ensure \textit{high performance} (e.g., high planning success rate and shorter moving times) and \textit{efficiency} (e.g., real-time planning and lower energy consumption). The perception module uses depth cameras to collect point clouds, which serve as input for the 3D semantic scene completion network \cite{jmwang} to predict and complete the lightly occluded environments, such as buildings or low bushes. This process generates a complete local map, which is used to construct the Euclidean Signed Distance Field (ESDF) map (Fig.~\ref{fig:head}a) for the planner to search for a collision-free trajectory.

Unfortunately, while these ESDF-based AGR navigation systems have proven successful in lightly occluded scenarios \cite{jmwang}, they face two limitations when navigating in cluttered environments with severe occlusions (e.g., forests). 

Firstly, the \textit{perception module} generates incomplete local maps in cluttered environments due to the low prediction accuracy of lightweight 3D semantic occupancy networks like SCONet from AGRNav\cite{jmwang}, resulting in high-risk collision paths (e.g., {\textit{red path}} in Fig.~\ref{fig:head}a). While employing transformers \cite{li2023voxformer} or 3D CNNs \cite{xia2023scpnet,cao2022monoscene} can improve accuracy, they are impractical for resource-constrained AGRs. Conversely, lightweight designs offer computational efficiency but compromise on precision.

Secondly, the existing ESDF-based AGR \textit{path planners} \cite{zhang2022autonomous,jmwang} are inefficient since building the ESDF map generates redundant calculation times that do not meet the real-time requirements. Furthermore, while \textit{Zhou et al.} \cite{zhou2020ego} devised an ESDF-free path planner for quadcopters, it fails to address AGR-specific requirements, particularly energy efficiency and dynamic constraints. Specifically, their flight-centric trajectory generation (e.g., {\textit{black path}} in Fig.\ref{fig:head}a.) results in elevated energy consumption and the inherent non-holonomic constraints of AGRs make it impossible to naively migrate and use such planners (Table \ref{label:1}). Notably, existing AGR path planners' inefficiencies stem from both the above inherent flaws and incomplete local maps provided by the perception module, leading to either overly conservative or aggressive mode switching, such as the choice of landing point location depicted in Fig.\ref{fig:head}a, which undermines energy efficiency.

Our key insight for addressing the limitations of the \textit{perception module} is to introduce a novel lightweight SSC network. This network aims to separate the conventional SSC network \cite{roldao2020lmscnet,jmwang}, which typically learns semantics and geometry together, into distinct network branches, thereby decoupling the processes that were previously jointly learned in networks. This separation enables each branch to focus on acquiring domain-specific features, thereby enhancing the overall performance. Concurrently, drawing inspiration from \cite{liu2023bevfusion,yang2021semantic,mei2023ssc}, we transition the feature fusion process to the Bird's Eye View (BEV) space, which holds the potential to diminish computational complexity and ensure high-speed inference, culminating in the generation of a complete local map.

Building upon the complete map produced by our SSC network, we next tackle the \textit{path planner's} limitations to ensure energy-efficient and real-time planning (i.e., meet the efficiency metrics in Table \ref{label:1}).  To achieve these, the path planner must accommodate the non-holonomic constraints inherent to AGRs and remove redundant ESDF calculations. Additionally, incorporating energy costs associated with different modes (e.g., flying and driving) is imperative for facilitating judicious mode switching and promoting energy conservation. 

\begin{figure*}[t]
  \centering
     \includegraphics[width=0.82\linewidth]{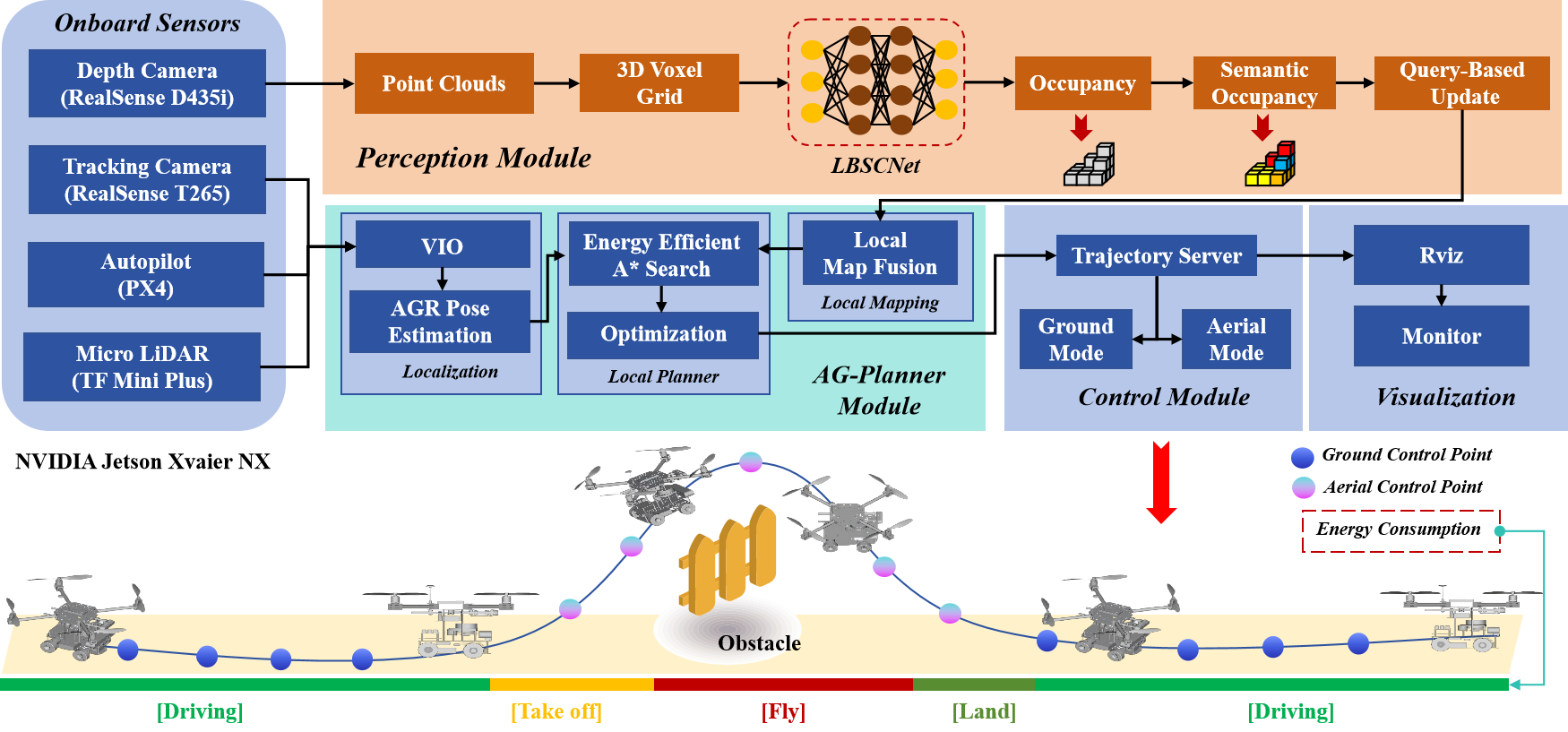}
   \caption{\small HE-Nav system architecture. The perception module and path planner run asynchronously on the onboard computer, connected through a query-based map update method \cite{jmwang} to ensure real-time local map updates with prediction results. }
   \label{fig:overview}
\end{figure*}

Based on these above insights, we present \textbf{\textit{HE-Nav}}, the first \underline{\textit{high-performance}} and \underline{\textit{efficient}} navigation system tailored for AGRs, as illustrated in Fig.~\ref{fig:overview}. The system comprises two key components, with the first being a lightweight BEV-guided semantic scene completion network (LBSCNet) deployed on the AGR. By processing sparse point cloud inputs, LBSCNet performs fast inference to accurately predict obstacle distribution (i.e., voxel occupancy) and semantics. These predictions are then integrated into local maps for path planning, facilitated by the query-based low-latency map update method presented in \cite{jmwang}, ensuring timely updates. 

During the planning phase, we develop an AG-Planner that searches for aerial-ground hybrid paths. Specifically, an energy-efficient kinodynamic A* path searching front-end utilizes motion primitives instead of straight lines as graph edges, by adding additional energy costs for aerial destinations, the planner not only tends to search ground trajectories but also switches to aerial mode only when AGRs encounter huge obstacles,  thereby promoting energy-saving. We then utilize a distance estimation method from \cite{zhou2020ego} to circumvent obstacles, avoiding ESDF computations. Finally, a gradient-based B-spline optimizer refines paths to generate a safe, smooth, and dynamically feasible trajectory.

We evaluated LBSCNet on the SemanticKITTI benchmark and compared its performance to some leading SSC networks. Then, we tested HE-Nav against two AGR navigation baselines (i.e., TABV \cite{zhang2022autonomous} and AGRNav \cite{jmwang}) in simulated and real environments, demonstrating its improved performance and efficiency (Table \ref{label:1}). Our evaluation reveals:
\begin{itemize}
\item \textbf{HE-Nav is high-performance.} HE-Nav achieved success rates of 98\% in the two simulation scenarios, while having the shortest average movement time.  (§~\ref{sec:C1})
\item \textbf{HE-Nav is efficient.} HE-Nav achieves \textbf{7x} energy savings in real-world tests, while reducing planning time by \textbf{6x} relative to ESDF-based baselines. (§~\ref{sec:D1})
\item \textbf{LBSCNet is accurate and high-speed inference.} LBSCNet achieves state-of-the-art performance (IoU = 59.71) on the SemanticKITTI benchmark and enables high-speed inference (i.e., 20.08 FPS). (§~\ref{sec:B1})
\end{itemize}

Our main contributions are the creation of the lightweight LBSCNet and the energy-efficient AG-Planner. \textbf{(1)} LBSCNet featuring a novel BEV fusion branch and SCB-Fusion module for fast inference and complete local map generation. \textbf{(2)} Leveraging this map, AG-Planner achieves energy-efficient, ESDF-free path planning.

\begin{table}[t]
\centering
\caption{Compared with three baseline AGR navigation systems and EGO-Planner designed specifically for multicopters.}
\resizebox{\columnwidth}{!}{%
\begin{tabular}{@{}lccccccc@{}}
\toprule
\multirow{2}{*}{Method} & \multirow{2}{*}{\scriptsize Suitable to AGRs} & \multirow{2}{*}{\scriptsize Category} & \multirow{2}{*}{\scriptsize Occ. Aware.} & \multicolumn{2}{c}{\scriptsize Perf. Metric} & \multicolumn{2}{c}{\scriptsize Eff. Metric} \\ \cmidrule(l){5-8} 
                        &                                   &                           &                                      & {\scriptsize Mov. Time} & {\scriptsize Succ. Rate} & {\scriptsize Plan. Time} & {\scriptsize Energy Cons.} \\ \midrule
{\scriptsize HDF \cite{fan2019autonomous}}  & \scriptsize\Checkmark  & {\scriptsize Only A*}  & \scriptsize\XSolidBrush   & \scriptsize\XSolidBrush    & \scriptsize\XSolidBrush   & \scriptsize\XSolidBrush  & \scriptsize\XSolidBrush    \\
{\scriptsize TABV \cite{zhang2022autonomous}}  & \scriptsize\Checkmark   & {\scriptsize ESDF-based}     & \scriptsize\XSolidBrush    & \scriptsize\XSolidBrush  & \scriptsize\XSolidBrush    & \scriptsize\XSolidBrush  & \scriptsize\XSolidBrush   \\
{\scriptsize AGRNav\cite{jmwang}}  & \scriptsize\Checkmark   & {\scriptsize ESDF-based}   & \scriptsize\Checkmark     & \scriptsize\XSolidBrush     & \scriptsize\XSolidBrush    & \scriptsize\XSolidBrush        & \scriptsize\XSolidBrush       \\
{\scriptsize EGO-Planner \cite{zhou2020ego}}   & \scriptsize\XSolidBrush      & {\scriptsize ESDF-free}     & \scriptsize\XSolidBrush    & \scriptsize\Checkmark & \scriptsize\Checkmark    & \scriptsize\Checkmark     & \scriptsize\XSolidBrush    \\
\midrule 
{\scriptsize \textbf{HE-Nav (Ours)}}    & \scriptsize\Checkmark       & {\scriptsize ESDF-free}      & \scriptsize\Checkmark      & \scriptsize\Checkmark    & \scriptsize\Checkmark  & \scriptsize\Checkmark   & \scriptsize\Checkmark    \\
\bottomrule
 \label{label:1}
\end{tabular}%
}
\end{table}

\section{RELATED WORKS}
\subsection{Motion Planning for AGRs}
Numerous researchers have explored various aerial-ground robot configurations, such as incorporating passive wheels \cite{zhang2022autonomous}, or multi-limb \cite{martynov2023morphogear} onto drones. In contrast, others \cite{tan2021multimodal,zhang2022coupled} have integrated rotors with wheeled robots to achieve dual-mode (i.e., flying and driving) locomotion. These designs facilitate enhanced stability and control in both locomotion modes. Consequently, we also adopted this mechanical structure to customize further our AGR, which has four wheels and four rotors. Moreover, Existing research primarily focuses on innovative mechanical structure designs \cite{zhang2022multi,wang2023design}, and the area of AGR autonomous navigation remains underexplored. Recently, \cite{fan2019autonomous} tackled ground-aerial motion planning, utilizing the A* algorithm for geometric path guidance and favouring ground paths by adding extra energy costs to aerial paths. However, this approach is limited by its lack of dynamic models and post-refinement in local planner trajectories, potentially compromising smoothness and dynamic feasibility. \cite{zhang2022autonomous} introduced an efficient and adaptive path planner and controller, but its reliance on an ESDF map results in intensive computation and limited perception of occluded areas, consequently leading to a low success rate in path planning and increased energy consumption. 

\subsection{Occlusion-Aware for AGRs}
AGR's sensor-based perception method cannot include the distribution of obstacles in the occluded area in the local map, leading to sub-optimal planned paths. Recent advancements in semantic scene completion \cite{xia2023scpnet} have addressed this challenge in complex and unknown environments, with the development of point-cloud-based \cite{yang2021semantic,mei2023ssc} and camera-based methods. \textit{Cao et al.} \cite{cao2022monoscene} introduced MonoScene, a camera-based approach that infers scene structure and semantics from a single monocular RGB image. In point-cloud-based methods, \cite{roldao2020lmscnet} introduced LMSCNet, a multiscale 3D semantic scene completion approach using a 2D UNet backbone with multiscale skip connections and a 3D segmentation head. However, the high computational demands of these methods limit their suitability for resource-constrained AGR platforms.

\section{Perception Module of HE-Nav}
\label{sec:3}
In this section, we introduce a lightweight three-branch SSC network (LBSCNet), depicted in Fig.~\ref{fig:lbscnet}. LBSCNet consists of a semantic branch, a completion branch, and a BEV fusion branch, serving as an alternative to conventional memory-intensive SSC networks \cite{cao2022monoscene,li2023voxformer,roldao2020lmscnet} that jointly predict geometry and semantics. By employing a pre-trained model offline on our AGR device (e.g., Jetson Xavier NX), LBSCNet can infer and predict the obstacle distribution in occluded areas at high speed. Subsequently, these prediction results are updated into a local map, which is utilized for path planning.

\subsection{LBSCNet Network Structure}

\begin{figure}[t]
  \centering
     \includegraphics[width=\linewidth]{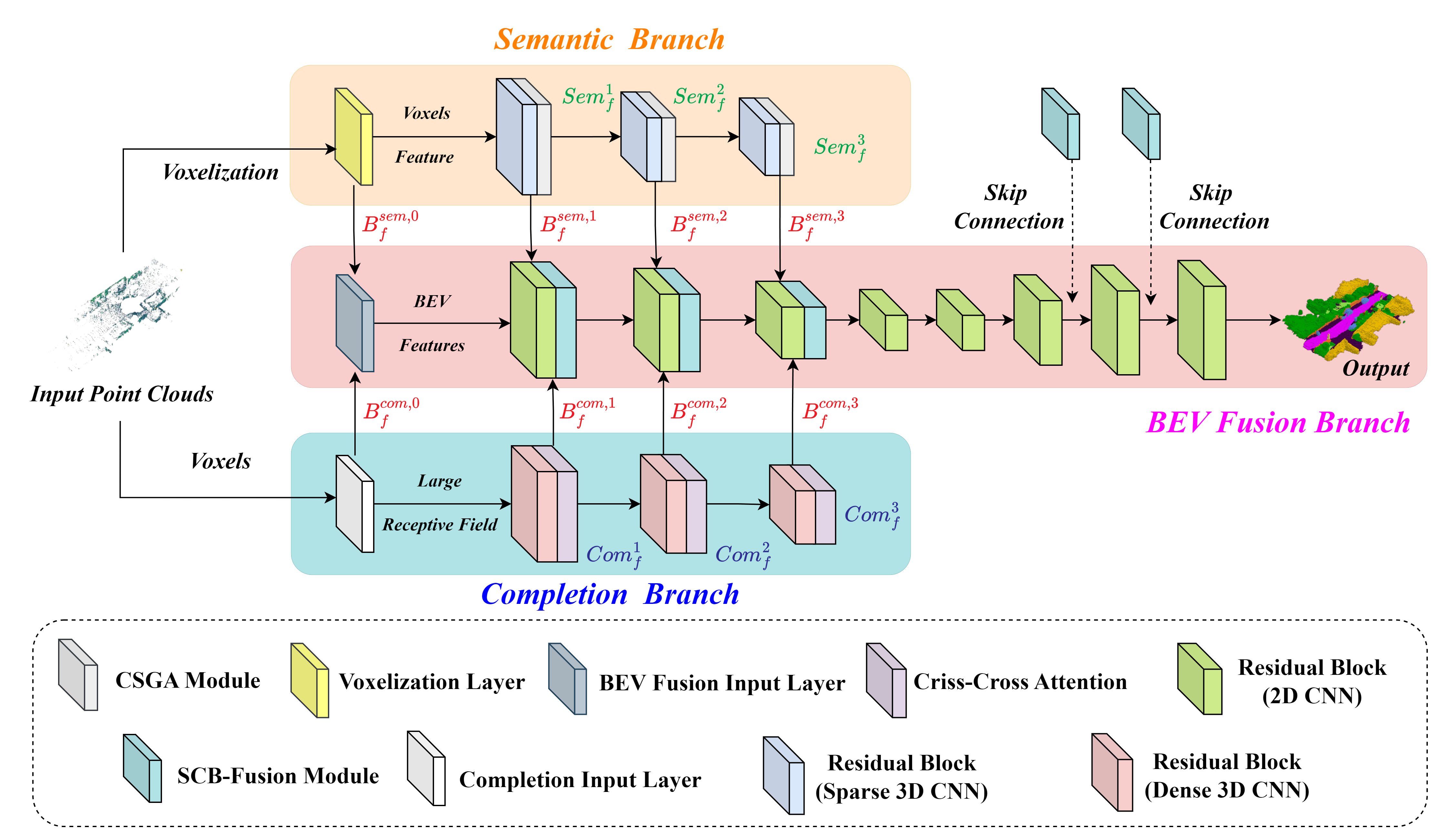}
   \caption{\small The overview of the proposed LBSCNet. It consists of semantic, completion and BEV fusion branches. }
   \label{fig:lbscnet}
\end{figure}

LBSCNet decoupling the learning process of semantics and completion, allows the network to concentrate on specific features (i.e., semantics and geometry), resulting in more efficient learning. The specific structures are as follows:

\noindent\textbf{\textit{Semantic Branch}:} This branch consists of a voxelization layer and three encoder blocks sharing a similar architecture, each encoder block comprises a residual block \cite{he2016deep} with sparse 3D convolutions and a cross-scale global attention (CSGA) module from \cite{mei2023ssc,xu2022sparse}. The integration of the CSGA module not only aligns multi-scale features with global voxel-encoded attention to capturing the long-range relationship of context but also alleviates the computational burden by reducing feature resolution. Specifically, in the voxelization layer, point clouds $P \in \mathbb{R}^{N \times 3}$ are partitioned based on the voxel resolution $s$ and mapped into voxel space. Subsequently, an aggregation function (i.e., max function) is applied to the point cloud within each voxel, yielding a single feature vector. A multi-layer perceptron (MLP) reduces the dimensionality of this feature vector, producing the final voxel features ${V}_{f_m}$ with a spatial resolution of $L \times W \times H$, $f_m$ represents the index of the voxel. The voxel features ${V}_{f_m}$ are then input into three encoder blocks to obtain semantic features $\{Sem_f^1, Sem_f^2, Sem_f^3\}$ (Fig.~\ref{fig:lbscnet}). The semantic branch is optimized using lovasz loss \cite{berman2018lovasz} and cross-entropy loss \cite{zhang2018generalized}. The semantic loss $L_{sem}$ is the sum of the loss at each stage, expressed as follows:
\begin{equation}
 L_{sem}=\sum_{i=1}^{3}(L_{cross,i} + L_{lovasz,i}) 
\end{equation}

.
\noindent\textbf{\textit{Completion Branch}:} The input to the completion branch is voxels $\small V \in \mathbb{R}^{1 \times L \times W \times H}$ generated by point clouds. The output is the multi-scale dense completion features $ \small \{Com_f^1, Com_f^2, Com_f^3\}$, providing more intricate geometric information. As depicted in Fig.~\ref{fig:lbscnet}, the completion branch comprises an input layer (kernel size $\small  7 \times 7 \times 7$), three residual blocks and three GPU memory-efficient criss-cross attention (CCA) \cite{huang2019ccnet} modules. The residual blocks incorporate dense 3D convolutions with a kernel size of $\small  3 \times 3 \times 3$, capturing local geometric features. Conversely, the criss-cross attention (CCA) \cite{huang2019ccnet} module is designed to capture long-range dependencies by gathering contextual information in both horizontal and vertical directions, thereby enriching the completion features with a global context.  The training loss $L_{com}$ for this branch is calculated as follows:
\begin{equation}
  L_{com}=\sum_{i=1}^{3}(L_{binary\_cross,i} + L_{lovasz,i}) 
\end{equation}
where $i$ denotes the $i-th$ stage of the completion branch and $L_{binary\_cross}$ indicates the binary cross-entropy loss. 

Notably, lightweight MLPs serve as auxiliary heads during training, attached after each encoder block in the semantic and completion branches for voxel predictions. At the inference stage, these heads are detached to preserve a streamlined network architecture.

\noindent\textbf{\textit{BEV Feature Fusion Branch}:} Previous research on SSC tasks has relied on fusing dense 3D features, resulting in considerable computational overhead and hindering deployment on resource-constrained AGR devices. We propose a lightweight BEV fusion branch specifically designed for SSC tasks, capitalizing on recent advancements in BEV perception \cite{liu2023bevfusion,mei2023ssc}. By projecting learned semantic and geometric features into BEV space and incorporating the innovative SCB-Fusion module, we significantly reduce computational demands while maintaining rapid inference capabilities. Specifically, our BEV fusion network employs a U-Net architecture from \cite{mei2023ssc} with 2D convolutions, featuring an input layer and four residual blocks in the encoder (Fig.~\ref{fig:lbscnet}). The process of projecting semantic and geometric features to BEV space is as follows: 

\noindent\textbf{\textit{Semantic Feature Projection}:} To project three-dimensional semantic features $\small  \{Sem_f^1, Sem_f^2, Sem_f^3\}$ into the two-dimensional BEV space, we first generate a BEV index based on the voxel index $f_m$ and then the features sharing identical BEV indices are aggregated using an aggregation function (e.g., the max function) to yield sparse BEV features. Utilizing the feature densification function offered by spconv \cite{spconv2022}, we generate dense BEV features $ \small  \{B_{f}^{sem,0}, B_{f}^{sem,1}, B_{f}^{sem,2}, B_{f}^{sem,3}\}$ based on the BEV index and sparse BEV features. 

\noindent\textbf{\textit{Geometric Feature Projection}:} For geometric features $\small \{Com_f^1, Com_f^2, Com_f^3\}$, we stack dense 3D features along the $z$-axis and apply 2D convolution to reduce the feature dimension, generating dense BEV features $\small \{B_{f}^{com,0}, B_{f}^{com,1}, B_{f}^{com,2}, B_{f}^{com,3}\}$. Subsequently, the projected features are input into the BEV fusion network (Fig.~\ref{fig:lbscnet}). The BEV loss
$L_{bev}$ is :
\begin{equation}
 L_{bev}=3\times  (L_{cross} + L_{lovasz})
\end{equation}

\begin{figure}[htp]
  \centering
     \includegraphics[width=0.4\linewidth]{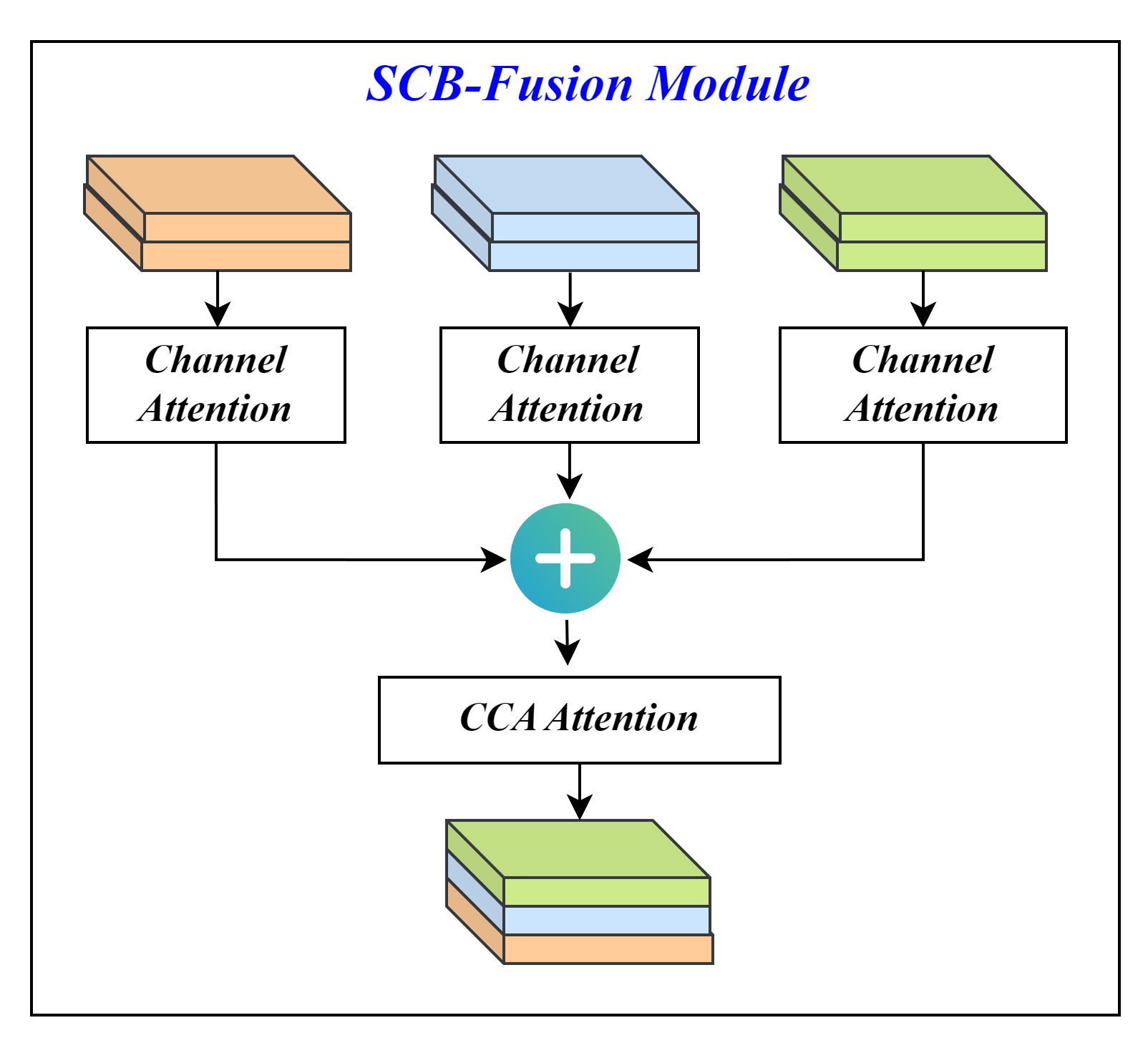}
   \caption{\small SCB-Fusion Module realizes the fusion of semantic features, geometric features and BEV features. }
   \label{fig:scb}
\end{figure}

\begin{figure*}[!t]
  \centering
   \begin{minipage}[b]{0.32\linewidth}
    \includegraphics[width=\linewidth,height=0.7\linewidth]{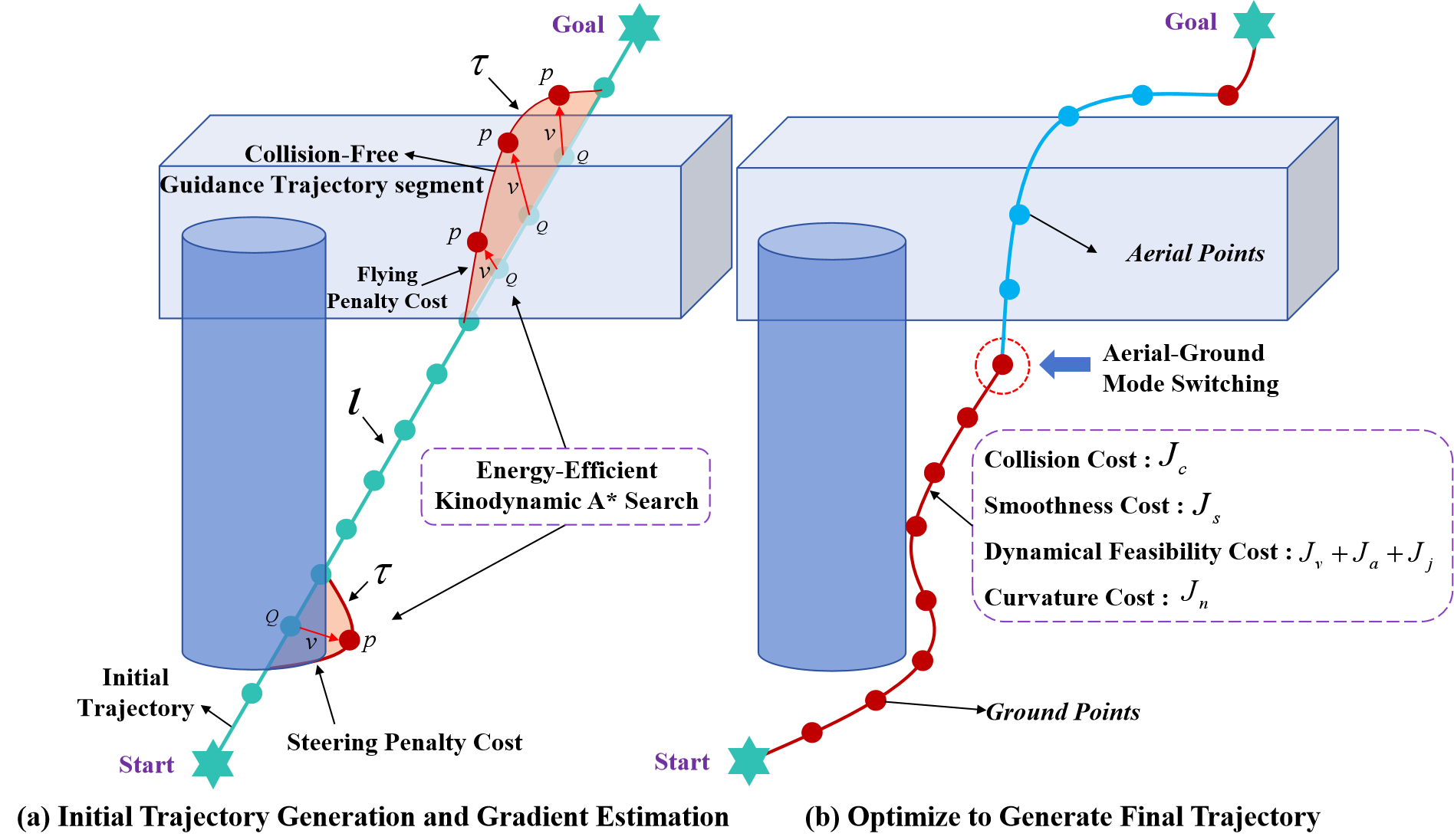}
    \caption{\small Illustration of AG-Planner and topological trajectory generation.}
    \label{fig:planner}
  \end{minipage}
  \hfill
  \begin{minipage}[b]{0.32\linewidth}
    \includegraphics[width=\linewidth]{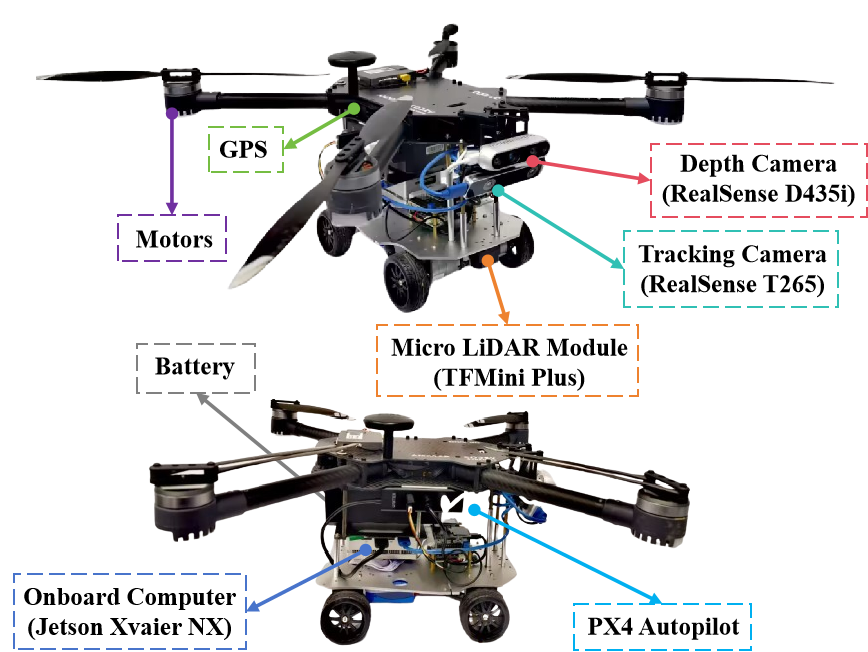}
    \caption{\small The detailed composition of our robot platform.}
    \label{fig:AGR}
  \end{minipage}
  \hfill
  \begin{minipage}[b]{0.32\linewidth}
    \includegraphics[width=\linewidth]{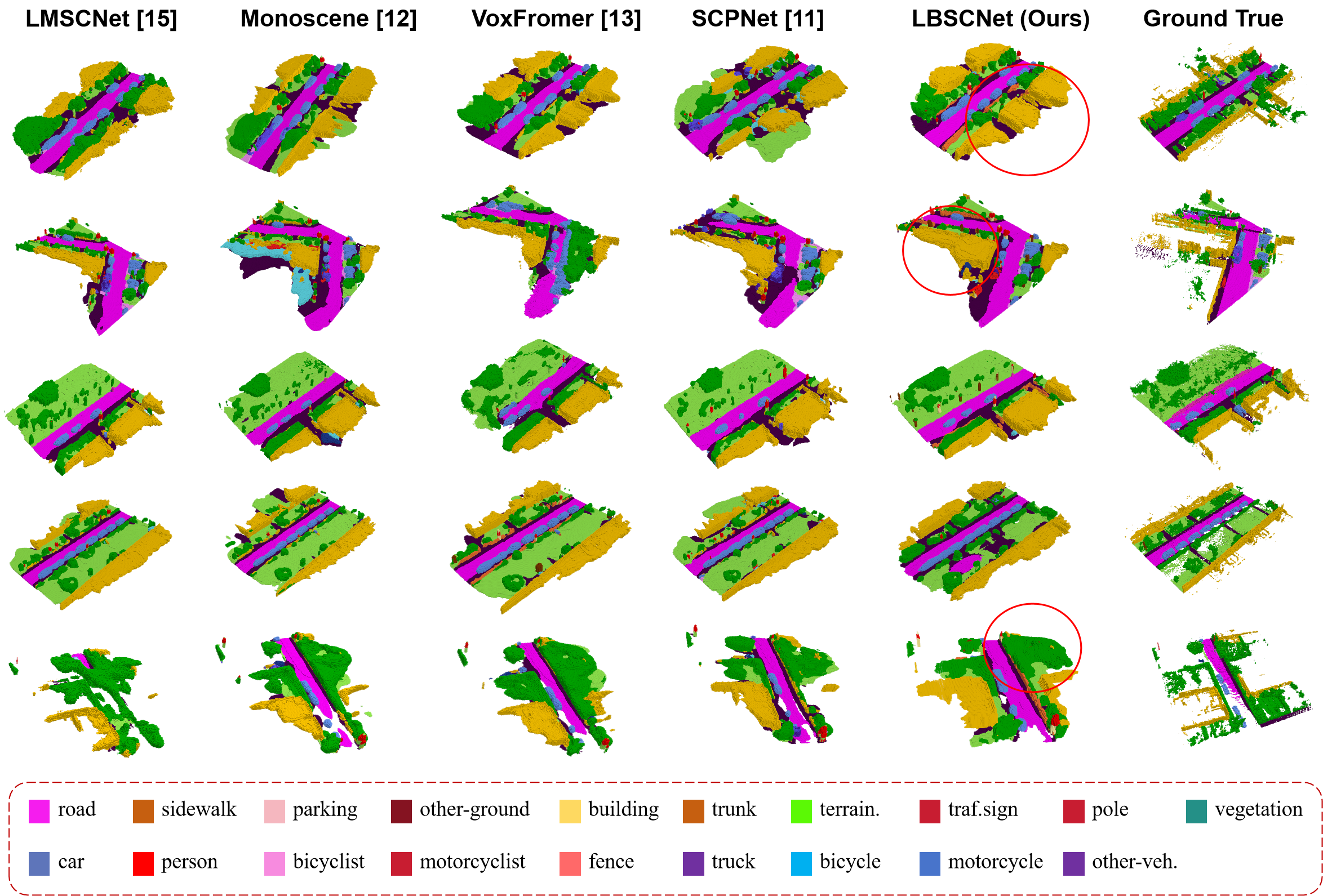}
    \caption{\small Qualitative results of our LBSCNet and others.}
    \label{fig:ssc}
  \end{minipage}
\end{figure*}

\noindent\textbf{\textit{Feature Fusion after Projection}:} To fuse the projected features, we devise an SCB-Fusion module (Fig.~\ref{fig:scb}) that fuses current semantic features, geometric features, and features from the previous layer. Specifically, we first compute channel attention for features $B_{pre}/B_{com}/B_{sem}$ to adaptively weight the feature channels. The weighted features are then summed and passed through a $1\times1$ convolution and CCA attention to obtain the fused features.

\noindent\textbf{\textit{LBSCNet Total Loss Function}:} The multi-task loss $L_{total}$ is expressed as :
\begin{equation}
 L_{total}=L_{bev} +  L_{sem} + L_{com}
\end{equation}
where $L_{bev}$, $ L_{sem}$ and $L_{com}$ respectively represent BEV loss, the semantic loss and completion loss.

\section{Aerial-Ground Motion Planner of HE-Nav}
\label{sec:4}
In this section, we introduce the novel AG-Planner. It is built on EGO-Planner \cite{zhou2020ego} and consists of \textbf{1)}  an energy-efficient kinodynamic A* path searching front-end,  \textbf{2)} a gradient-based trajectory optimization back-end to ensure ESDF-free and \textbf{3)} a post-refinement procedure.

\subsection{Energy-Efficient Kinodynamic Hybrid A* Path Searching}

Our AG-Planner first creates a naive ``\textit{initial trajectory}" $\iota$  (in Fig.~\ref{fig:planner}a) that overlooks obstacles by randomly adding coordinate points, considering the positions of both the starting and target points. Following that, for the ``\textit{collision trajectory segment}" (i.e., the trajectory inside the obstacle),  the back end of our planner based on \cite{dolgov2008practical} to propose an energy-efficient kinodynamic A* path search algorithm to establish a safe ``\textit{guidance trajectory segment}"  $\tau $, which uses motion primitives instead of straight lines as graph edges in the searching loop. In this algorithm, we add extra flying costs for the motion primitives. Consequently, the path searching not only tends to plan ground trajectories but also switches to aerial mode and flies over them only when AGRs encounter huge obstacles,  thereby promoting energy-saving. 

\subsection{Gradient-Based B-spline Trajectory Optimization}
\noindent\textbf{\textit{B-spline Trajectory Formulation}:} In trajectory optimization (in Fig.~\ref{fig:planner}b),  the trajectory is parameterized by a uniform B-spline curve $\Theta$, which is uniquely determined by its degree $p_b$, $N_c$ control points $\left \{ Q_1,Q_2,Q_3,...,Q_{N_c} \right \} $, and a knot vector $\left \{ t_1,t_2,t_3,...,t_{M-1},t_M \right \} $, where $Q_i\in \mathbb{R}^3,t_m\in\mathbb{R},M=N+p_b$. Following the matrix representation of the \cite{boor1971subroutine} the value of a B-spline can be evaluated as:
\begin{equation}
 \Theta(u)=[1, u,...,u^p]\cdot M_{p_b+1} \cdot[Q_{i-p_b},Q_{i-p_b+1},...,Q_i]^T 
\end{equation}
where $M_{p_{b}+1}$ is a constant matrix depends only on $p_b$. And $\small u=(t-t_i)/(t_{i+1}-t_i)$, for $t\in [t_i,t_{i+1})$. 

In particular, in ground mode, we assume that AGR is driving on flat ground so that the vertical motion can be omitted and we only need to consider the control points in the two-dimensional horizontal plane, denoted as $Q_{ground}=\left \{ Q_{t0},Q_{t1},...,Q_{tM} \right \} $, where $Q_{ti}=(x_{ti},y_{ti}),i\in \left [ 0,M \right ]$. In aerial mode, the control points are denoted as $Q_{aerial}$. According to the properties of B-spline: the $k^{th}$ derivative of a B-spline is still a B-spline with order $p_{b,k}=p_b-k$, since $\bigtriangleup t$  is identical alone $\Theta$, the control points of the velocity $V_i$, acceleration $A_i$ and jerk $J_i$ curves are obtained by:
\begin{equation}
 V_i=\frac{Q_{i+1}-Q_i}{\bigtriangleup t},A_i=\frac{V_{i+1}-V_i}{\bigtriangleup t}, J_i=\frac{A_{i+1}-A_i}{\bigtriangleup t}  
\end{equation}

\noindent\textbf{\textit{Collision Avoidance Force Estimation}:} Different from ESDF-based methods \cite{zhang2022autonomous,jmwang}, for each control point on the collision trajectory segment, vector $v$ (i.e., a safe direction pointing from inside to outside of that obstacle) is generated from $\iota$ to $\tau$ and $p$ is defined at the obstacle surface (in Fig.~\ref{fig:planner}a). With generated $\left \{ p,v \right \} $ pairs, the planner maximizes $D_{ij}$ and returns an optimized trajectory. The obstacle distance $D_{ij}$ if $i^{th}$ control point $Q_i$ to $j^{th}$ obstacle is defined as:
\begin{equation}
D_{ij}=(Q_i-p_{ij})\times v_{ij}
\end{equation}
Because the guide path $\tau$ is energy-saving, the generated path is also energy efficient.

\noindent\textbf{\textit{B-spline Trajectory Optimization and Post-refinement Procedure}:} The basic requirements of the B-spline paths are three-fold: \textit{smoothness}, \textit{safety}, and \textit{dynamical feasibility}. Based on the special properties of AGR bimodal, we first adopt
the following cost terms designed by \textit{Zhou et al.} \cite{zhou2020ego}:
\begin{equation}
\min J_1= \lambda_sJ_s + \lambda_cJ_c + \lambda_f(J_v + J_a + J_j) 
\end{equation}
where $J_s$ is the smoothness penalty, $J_c$ is for collision, and $J_v, J_a, J_j$ are dynamical feasibility costs that limit
velocity, acceleration and jerk. $\lambda_s,\lambda_c,\lambda_f$ are weights for each cost terms. Detailed explanations can be found in \cite{zhou2020ego}. Subsequently, based on our observations, AGR faces non-holonomic constraints when driving on the ground, which means that the ground velocity vector of AGR must be aligned with its yaw angle. Additionally, AGR needs to deal with curvature limitations that arise due to minimizing tracking errors during sharp turns. Therefore, a cost for curvature needs to be added, and $J_n$ can be formulated as:
\begin{equation}
 J_n=\sum_{i=1}^{M-1}F_n(Q_{ti}) 
\end{equation}
where $F_n(Q_{ti})$  is a differentiable cost function with $C_{max}$ specifying the curvature threshold:
\begin{equation}
F_n(Q_{ti})=\left\{\begin{matrix} 
  (C_i-C_{max})^2,C_i>C_{max}, \\  
  0,C_i\le C_{max} 
\end{matrix}\right. 
\end{equation}
where $C_i=\frac{\bigtriangleup \beta _i}{\bigtriangleup Q_{ti}} $ is the curvature at $Q_{ti}$, and the $\small \bigtriangleup \beta _i=\left | \tan ^{-1} \frac{\bigtriangleup y_{ti+1}}{\bigtriangleup x_{ti+1}} - \tan ^{-1} \frac{\bigtriangleup y_{ti}}{\bigtriangleup x_{ti}}\right | $ . In general, the overall objective function is formulated as follows:
\begin{equation}
\min J_{all}= \lambda_sJ_s + \lambda_cJ_c + \lambda_f(J_v + J_a + J_j)  + \lambda_n J_n\\
\end{equation}

The optimization problem is solved using the non-linear optimization solver NLopt \cite{johnson2014nlopt}, with post-refinement from \cite{zhou2020ego} for constraint violations. After path planning, a setpoint from the trajectory is selected and sent to the controller. In addition, when the $z$-axis coordinate of the next control point is greater than the ground threshold, that is, when mode switching is required, an additional trigger signal will be sent to the controller (i.e., PX4 Autopilot). The controller will automatically switch to the flight state.

\section{Evaluation}
\label{sec:5}
In this section, we first assess the LBSCNet on the SemanticKITTI benchmark. Subsequently, we selected the model with the best completion accuracy on SemanticKITTI and integrating this model with AG-Planner, a comprehensive HE-Nav system is formed. We then evaluate the AGR's autonomous navigation capability using HE-Nav in both simulated and real-world settings, focusing on \textit{\textbf{performance}} metrics (i.e., planning success rate, average movement time) and \textit{\textbf{efficiency}} metrics (i.e., average planning time, energy consumption). Finally, ablation experiments verify the navigation performance and efficiency improvements brought by HE-Nav's two key components.

\subsection{Evaluation setup} 
\noindent\textbf{\textit{Perception Module:}} We trained and tested LBSCNet using the outdoor SemanticKITTI dataset \cite{geiger2012we} on a single NVIDIA 3090 GPU. The model was trained for 80 epochs with a batch size of 8, using the Adam optimizer \cite{kingma2014adam} at an initial learning rate of 0.001, and input point cloud augmentation by random flipping along the $x-y$ axis. Finally, we deployed the pre-trained model offline with the best completion accuracy to complete the local map.

\noindent\textbf{\textit{Simulation Experiment:}} The simulation environments included a densely cluttered $20m \times 20m \times 5m$ square room, featuring 80 walls and 20 rings, and a similarly cluttered $3m \times 30m \times 5m$ corridor, containing 60 walls and 10 rings, both of which presented significantly occluded spaces as depicted in Fig.~\ref{fig:sim2}. These scenarios incorporated 40\% more obstacles compared to our prior work, AGRNav \cite{jmwang}. In these simulations, the AGR's task was to navigate from a starting point to a designated destination without collision.

\noindent\textbf{\textit{Real-world Experiment:}} We employed HE-Nav on a custom AGR platform (Fig.~\ref{fig:AGR}) for indoor and outdoor experiments, using Prometheus software \cite{Prometheus} with a RealSense D435i depth camera, a T265 tracking camera, GPS \cite{wang2020application}, and a Jetson Xavier NX onboard computer. We recorded the average energy consumption per second for AGR during driving and flying (Table~\ref{tab:energy}) to establish a basis for evaluating energy usage in real and simulated tests.

\subsection{LBSCNet Comparison against the state-of-the-art.}
\label{sec:B1}
\noindent\textbf{\textit{Quantitative Results:}} We evaluated our proposed LBSCNet against state-of-the-art SSC methods on the SemanticKITTI test datasets by submitting results to the official test server. Table~\ref{tab:ssc} demonstrates that LBSCNet not only achieves the highest completion metric IoU (59.71\%) but also ranks third in the semantic segmentation metric mIoU (23.58\%). Although SCPNet's semantic segmentation accuracy surpasses ours, its dense network design renders it incapable of real-time inference. In contrast, LBSCNet outperforms SCPNet by 6.43\% in IoU and runs approximately \textbf{\textit{20 times}} faster.

\begin{table}[htp]
\begin{center}
\footnotesize
\caption{\small Quantitative comparison against the state-of-the-art SSC methods.}
\renewcommand{\arraystretch}{1.2}
\begin{tabularx}{\columnwidth}{@{}>{\centering\arraybackslash}p{2.8cm}*{5}{>{\centering\arraybackslash}X}@{}}  
\toprule
\textbf{Method} & \textbf{\textit{IoU}} & \textbf{\textit{mIoU}} & \textbf{\textit{Prec.}} & \textbf{\textit{Recall}} & \textbf{\textit{FPS}} \\
\midrule
SSCNet \cite{song2017semantic}  & 53.20 & 14.55 & 59.13 & \textbf{84.15} & 12.00 \\
LMSCNet \cite{roldao2020lmscnet} & 55.32 & 17.01 & 77.11 & 66.19 & 13.50 \\
LMSCNet-SS \cite{roldao2020lmscnet} & 56.72 & 17.62 & 81.55 & 65.07 & 13.50 \\
SCONet \cite{jmwang} & 56.12 & 17.61 & \textbf{85.02} & 63.47 & 20.00 \\
S3CNet \cite{cheng2021s3cnet} & 45.60 & 29.50 & 48.79 & 77.13 & 1.20 \\
Monoscene \cite{cao2022monoscene} & 38.55 & 12.22 & 51.96 & 59.91 & $< 1$ \\
VoxFromer-T \cite{li2023voxformer} & 57.69 & 18.42 & 69.95 & 76.70 & $< 1$ \\
VoxFromer-S \cite{li2023voxformer} & 57.54 & 16.48 & 70.85 & 75.39 & $< 1$ \\
SCPNet \cite{xia2023scpnet} & 56.10 & \textbf{36.70} & 72.43 & 78.61 & $< 1$ \\
\midrule
\textbf{LBSCNet (Ours)} & \textbf{59.71} & 23.58 & 77.60 & 71.29 & \textbf{20.08} \\
\bottomrule
\end{tabularx}
\label{tab:ssc}
\end{center}
\end{table}

\noindent\textbf{\textit{Qualitative Results:}} We provide visualization results on the SemanticKITTI validation set. As illustrated in Fig.~\ref{fig:ssc}, our LBSCNet demonstrates superior SSC predictions, particularly for ``wall" classes and larger objects like cars, aligning with the results in Table~\ref{tab:ssc}. Importantly, the occlusion areas we target, such as vegetation and trees behind walls, are accurately completed, proving vital for subsequent path-planning.

\begin{table}[htp]
\caption{\small Ablation study on the SemanticKITTI validation set.}
\footnotesize
\centering
\begin{tabular}{lll}
\toprule
Method & IoU $\uparrow$ & mIoU $\uparrow$ \\
\midrule
LBSCNet (ours) & 58.34 & 22.74 \\
w/o SCB-Fusion Module & 57.05 & 21.26 \\
w/o Criss-Cross Attention & 57.20 & 22.17 \\
\bottomrule
\end{tabular}
\label{tab:ablation}
\end{table}

\noindent\textbf{\textit{Ablation Study:}} Ablation studies conducted on the SemanticKITTI validation set (Table~\ref{tab:ablation}) emphasize the significance of two crucial components in our network: CCA attention mechanisms and the SCB-Fusion Module. The CCA attention mechanism greatly influences completion accuracy by effectively aggregating context across rows and columns. The absence of CCA results in a 1.95\% decrease in completion accuracy. On the other hand, the SCB-Fusion module captures local scene features, including occluded areas, with minimal computational overhead. Removing the SCB-Fusion module leads to a 2.21\% reduction in IoU.

\begin{table}[htp]
  \renewcommand{\arraystretch}{1.2} 
  \begin{center}
  \caption{Battery and Energy Consumption Parameters}
    \begin{tabular}{p{5cm}|p{3cm}}

      \hline
      \textbf{Parameter} & \textbf{Value} \\
      \hline
      
     Battery Capacity & 10000 mAh \\
     
     Battery Weight & 1008 g \\
     
      Rated Power & 231 Wh \\
     
      Operating Voltage & 23.05 V \\
      \hline
      Driving Energy Consumption & $\approx$ 251.45 J/s \\
     
      Flying Energy Consumption  & $\approx$ 988.33 J/s \\
      \hline
    \end{tabular}
    
    \label{tab:energy}
  \end{center}
\end{table}

\subsection{Simulated Air-Ground Robot Navigation} 
\label{sec:C1}
\begin{figure}[htp]
  \centering
     \includegraphics[width=0.9\linewidth]{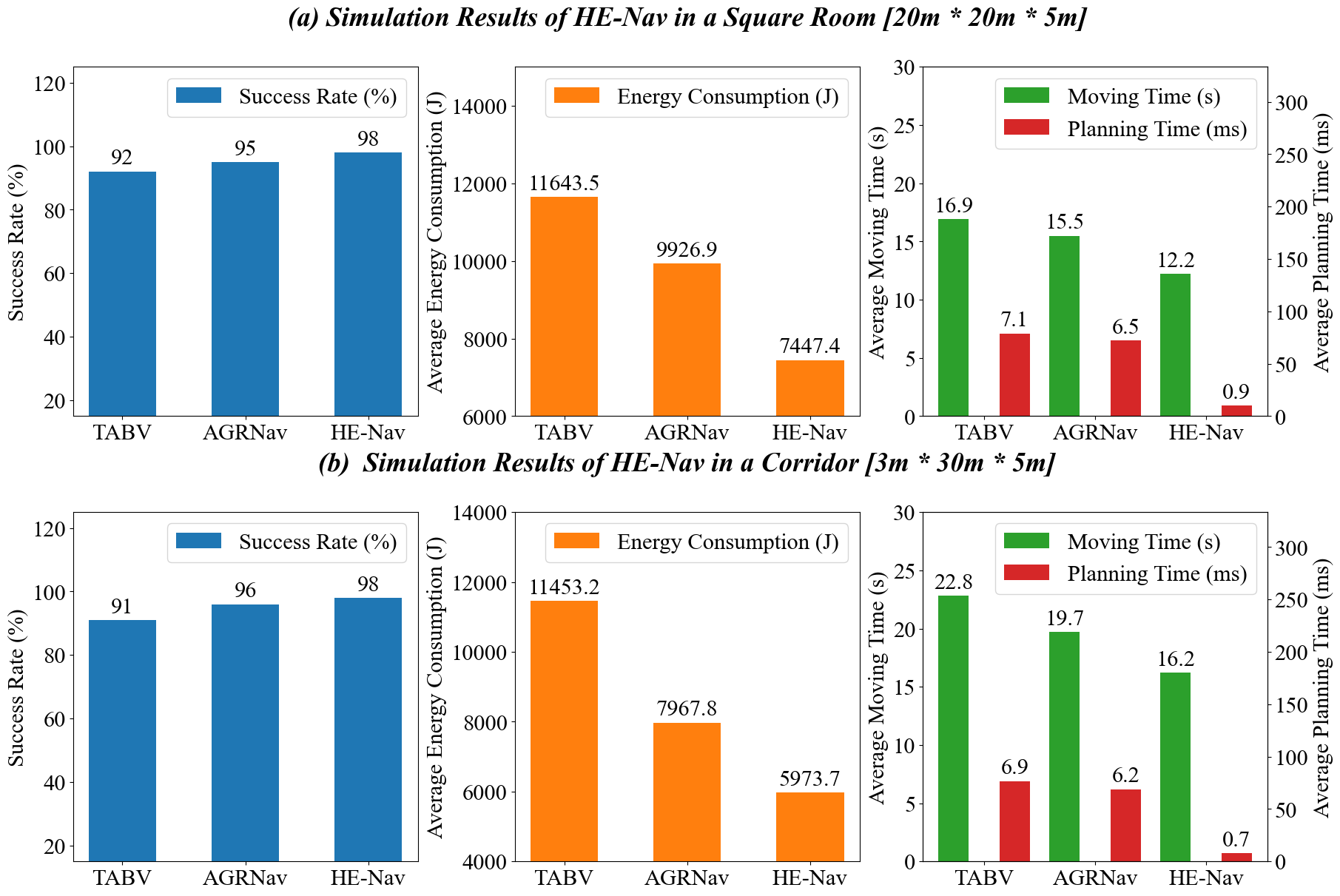}
   \caption{\small Quantitative results of HE-Nav in two simulation scenarios. }
   \label{fig:sim1}
\end{figure}

\begin{figure}[htp]
  \centering
     \includegraphics[width=0.8\linewidth]{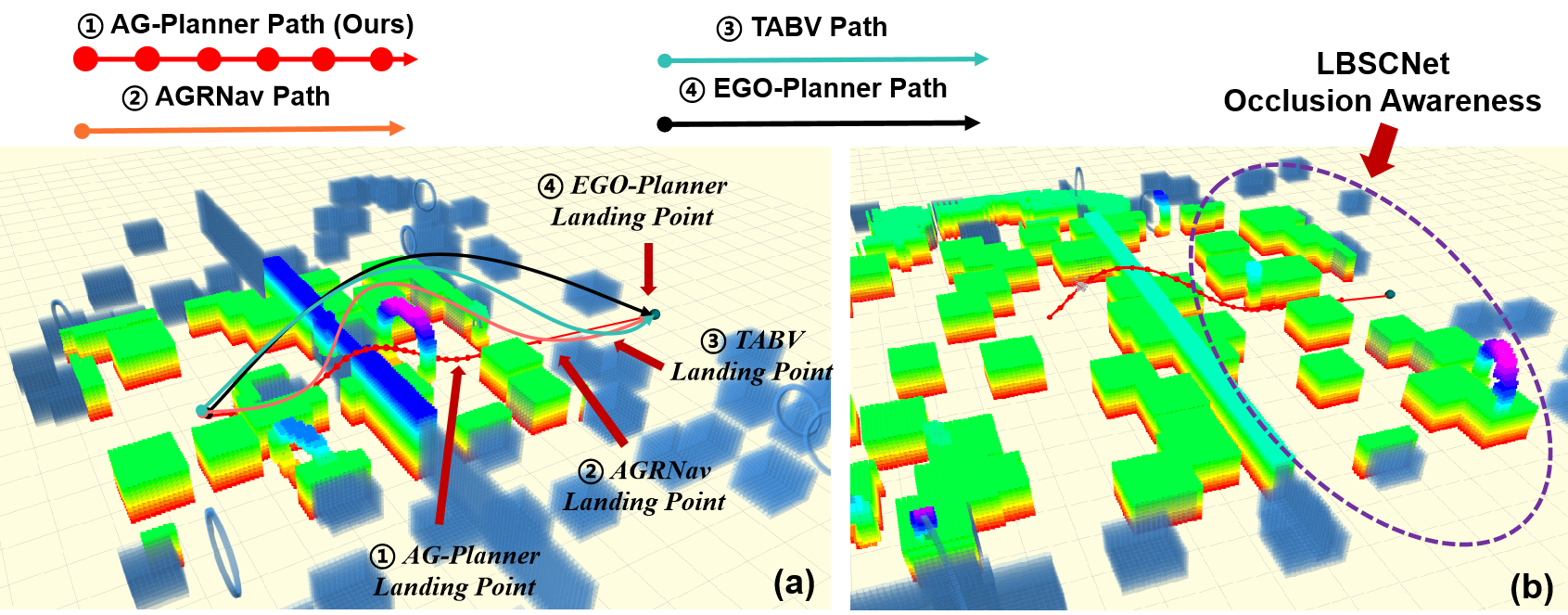}
   \caption{\small Qualitative results of path planning and occlusion prediction in simulation environment. }
   \label{fig:sim2}
\end{figure}

In a square room and corridor scenario (Fig~\ref{fig:sim2}), through 100 trials with varied obstacle placements, we evaluated the average moving time, planning time, and success rate (i.e., collision-free) of each method (Fig.~\ref{fig:sim1}). We additionally utilize the data from Table~\ref{tab:energy} and record the flight and driving durations (Table~\ref{tab:time}) in the simulation to compute the energy consumption.

\begin{table}[htp]
\centering
\caption{Flight and driving time and energy consumption results}
\label{tab:time}
\begin{scriptsize} 
\begin{tabular}{@{}lcccr@{}} 
\toprule
Method & Scene & Flying Time(s)  & Driving Time(s) & Total Energy(J) \\
\midrule
TABV \cite{zhang2022autonomous} & Sq. Rm. & 10.11  & 6.5678  & 11643.5 \\
AGRNav \cite{jmwang} & Sq. Rm. & 8.16  & 7.4055  & 9926.9 \\
\textbf{HE-Nav (ours)} & Sq. Rm. & \textbf{5.98} & \textbf{6.1132}  & \textbf{7447.4} \\
\midrule
TABV \cite{zhang2022autonomous} & Corr. & 7.74  & 15.1263  & 11453.2 \\
AGRNav \cite{jmwang} & Corr. & 4.06   & 15.7295  & 7967.8  \\
\textbf{HE-Nav (ours)} & Corr. & \textbf{2.54}  & \textbf{13.7734} & \textbf{5973.7} \\
\bottomrule
\end{tabular}
\end{scriptsize} 
\end{table}

Fig.~\ref{fig:sim1} showcases the exceptional performance of our HE-Nav system, achieving 98\% success rates in square rooms and corridors, with average movement times of 12.2s and 16.2s. Our system significantly accelerates planning time, being 6 times faster than TABV \cite{zhang2022autonomous} and AGRNav \cite{jmwang}, due to eliminating redundant ESDF calculations. Addressing the limitations of TABV, which lacks obstacle sensing in occluded areas, and AGRNav, with lower obstacle prediction accuracy, HE-Nav effectively employs the ESDF-free AG-Planner. Leveraging LBSCNet's precise obstacle prediction (Fig.~\ref{fig:sim2}b) and integrating the energy-efficient kinodynamic A* algorithm, our system achieves the lowest average energy consumption of 7447.4 J and 5973.7 J. AG-Planner also achieves the mode-switching balance between radical and conservative (e.g., optimal landing position in Fig.~\ref{fig:sim2}a), further enhancing energy savings.

\begin{table}[htp]
\centering
\caption{\small Ablation Study of HE-Nav Navigation }
\label{tab:ablation-experiments}
\begin{tabular}{@{}cccc@{}}
\toprule
Percep. & Plan. & Succ. Rate (\% )& Plan. Time (s) \\
\midrule
SCONet \cite{jmwang} & H-Planner \cite{jmwang}    & 95   &  6.5 \\
LBSCNet & H-Planner \cite{jmwang}    & 96   &  6.5 \\
SCONet \cite{jmwang} & AG-Planner   & 96 & 0.7 \\
\bottomrule
 - & AG-Planner & 95  & 0.7 \\
 LBSCNet & AG-Planner  & 98   & 0.7  \\
\bottomrule
 \label{tab:abl}
\end{tabular}
\end{table}

\begin{figure}[htp]
  \centering
     \includegraphics[width=0.7\linewidth]{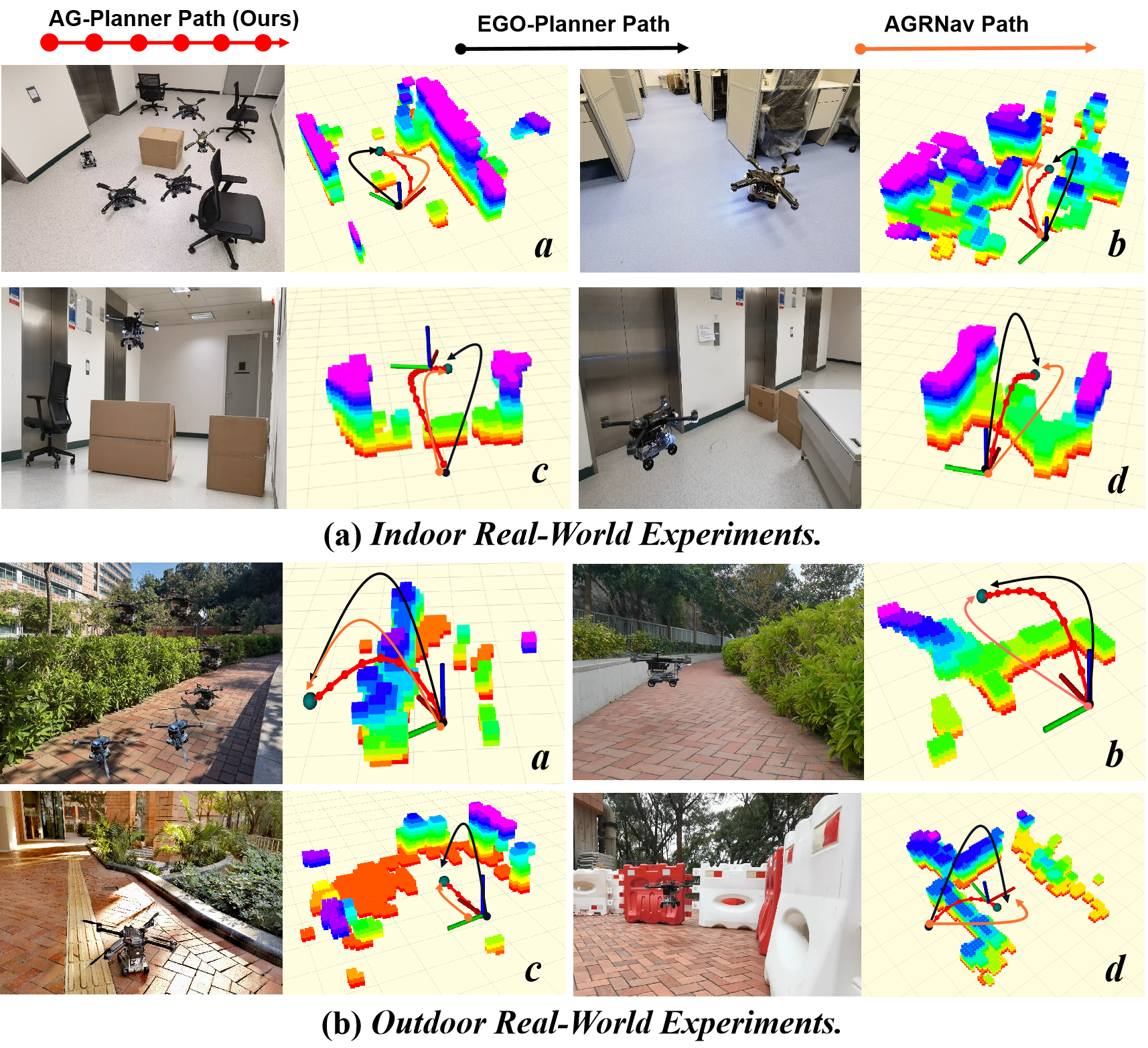}
   \caption{\small HE-Nav effectively predicts obstacle distribution in occluded areas and plans collision-free hybrid trajectories. }
   \label{fig:exp}
\end{figure}
\begin{figure}[htp]
  \centering
     \includegraphics[width=0.7\linewidth]{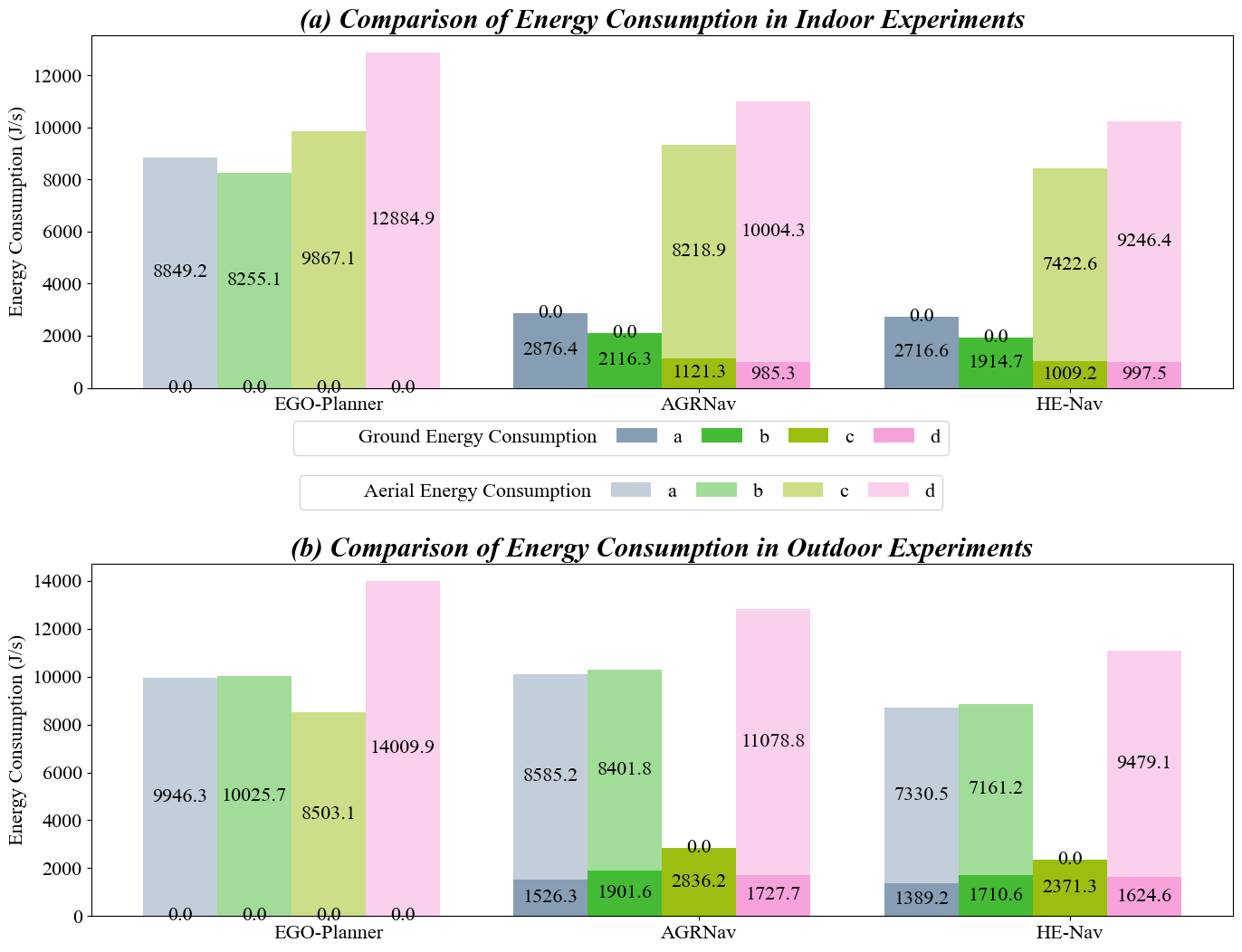}
   \caption{\small Quantitative results of indoor and outdoor real environmental energy consumption. }
   \label{fig:real-energy}
\end{figure}

\noindent\textbf{\textit{Ablation Study:}} In our ablation experiments (Table~\ref{tab:abl}), we conducted 100 trials in scenes containing 20 rings and 80 walls each, with obstacles placed differently in every trial, to validate the effectiveness of HE-Nav's components in cluttered occluded environments. AG-Planner alone achieved a 95\% success rate. Incorporating LBSCNet boosted the success rate by 3\%, while SCONet's addition only yielded a 1\% improvement. This disparity stems from SCONet's completion accuracy being 3.59\% lower than LBSCNet's, rendering it less efficient in handling complex occlusions. Moreover, AG-Planner reduces 87\% of computational redundancy, enabling millisecond-level path planning.

\subsection{Real-world Air-Ground Robot Navigation}
\label{sec:D1}
We assess HE-Nav's performance and energy efficiency across 4 indoor and 4 outdoor scenarios. In indoor settings (Fig.~\ref{fig:exp}(a)), our HE-Nav consistently demonstrates lower average energy consumption than AGRNav and EGO-Planner. For instance, in scenarios a and b, our system achieves energy consumption reductions of 69.3\% and 76.8\% relative to EGO-Planner (Fig.~\ref{fig:real-energy}), primarily due to the inclusion of additional penalty terms in the aerial segment, prompting our AG-Planner to favour energy-saving ground paths. Simultaneously, LBSCNet rapidly predicts obstacle distribution in occluded areas, generating a more comprehensive local map (e.g., b visualization results) to serve as the basis for AG-Planner's path search. This high-precision completion aids the planner in identifying optimal landing points, further contributing to energy conservation.

Transitioning to outdoor scenarios (Fig.~\ref{fig:exp}(b)), HE-Nav surpasses AGRNav with a 13.29\% reduction in average energy consumption in scenario d, This can be attributed not only to the optimization of smooth aerial paths, which minimizes flight energy consumption but also to LBSCNet's ability to accurately predict obstacle distribution in occluded environments (i.e., Fig.~\ref{fig:exp}(a)-b). This ability aids in reducing redundant paths and identifying optimal mode-switching points (e.g., the landing point in Fig.~\ref{fig:exp}(b)-d), as early landing and transition to ground driving mode promote energy conservation.

\section{CONCLUSION}
\label{sec:6}
We have presented HE-Nav, the first high-performance, efficient and ESDF-free navigation system specifically designed for aerial-ground robots (AGRs). By integrating key components such as the lightweight BEV-guided semantic scene completion network (LBSCNet) and the aerial-ground motion planner (AG-planner), our system is capable of predicting obstacle distributions in occluded areas and generating low-collision risk, energy-efficient trajectories in real-time ($\approx$ 1 ms). Through extensive simulations and real experiments, HE-Nav has been shown to significantly outperform recent AGR navigation systems and muticopter path planners in performance and efficiency.

\section{Acknowledgments}

The work is supported in part by National Key R\&D Program of China (2022ZD0160201), HK RGC RIF (R7030-22), HK ITF (GHP/169/20SZ), a Huawei Flagship Research Grant in 2023, HK RGC GRF (Ref: 17208223 \& 17204424), and the HKU-CAS Joint Laboratory for Intelligent System Software.


\bibliographystyle{IEEEtran}
\bibliography{refs}

\vfill

\end{document}